\documentclass[11pt]{article}

\usepackage[preprint]{acl}

\usepackage{times}
\usepackage{latexsym}
\usepackage{amsfonts}
\usepackage{booktabs}
\usepackage{amsmath}

\usepackage[T1]{fontenc}

\usepackage[utf8]{inputenc}

\usepackage{microtype}

\usepackage{inconsolata}

\usepackage{graphicx}

\title{Overalignment in Frontier LLMs: An Empirical Study of Sycophantic Behaviour in Healthcare}

\author{Cl{\'e}ment Christophe \\\And
  Wadood Mohammed Abdul \\\And
  Prateek Munjal \\\AND
  Tathagata Raha \\\And
  Ronnie Rajan \\\And
  Praveenkumar Kanithi \\\AND
  \textbf{M42, Abu Dhabi}}

\begin{document}
\maketitle
\begin{abstract}
As LLMs are increasingly integrated into clinical workflows, their tendency for \textit{sycophancy}, prioritizing user agreement over factual accuracy, poses significant risks to patient safety. While existing evaluations often rely on subjective datasets, we introduce a robust framework grounded in medical MCQA with verifiable ground truths. We propose the \textbf{Adjusted Sycophancy Score ($S_a$)}, a novel metric that isolates alignment bias by accounting for stochastic model instability, or ``confusability.'' Through an extensive scaling analysis of the Qwen-3 and Llama-3 families, we identify a clear scaling trajectory for resilience. Furthermore, we reveal a counter-intuitive vulnerability in reasoning-optimized ``Thinking'' models: while they demonstrate high vanilla accuracy, their internal reasoning traces frequently rationalize incorrect user suggestions under authoritative pressure. Our results across frontier models suggest that benchmark performance is not a proxy for clinical reliability, and that simplified reasoning structures may offer superior robustness against expert-driven sycophancy.
\end{abstract}

\section{Introduction}
Large Language Models (LLMs) have demonstrated remarkable capabilities across diverse domains using a mixture of different training techniques such as supervised finetuning (SFT), followed by Reinforcement learning (RL) techniques including Human/AI feedback, and verifiable rewards. Recently, RL based methods are employed to align LLMs with human intention. Although it has shown to improve adherence to user intentions by making the LLMs more helpful, they are also noted to lead to a behavior known as \textit{sycophancy}. Under this behavior, the model tends to strongly align its responses with the user's stated views or misconceptions, even when they contradict established facts \cite{casper2023open}.

In creative or open-ended tasks, this behavior/trait may be perceived as a form of "user-alignment". However, we note that in high stakes domains such as healthcare, sycophancy poses a serious safety risks and represent a major blocker in clinical adoption of LLMs. We hypothesize that an ideal clinical LLM must consistently prioritize medical knowledge/reasoning over user preferences, especially in clinical decision settings. 



In this paper, we address the critical gap in clinical AI safety by developing a robust sycophancy evaluation framework grounded in Medical MCQA benchmarks \citep{jin2020disease, wang2024mmlu}. By using exams with verifiable ground truths, we provide an objective measure of a model's resilience against explicit misinformation , presented in inputs via nudges/perturbations. 

Our contributions are three-fold. First, we introduce the \textbf{Adjusted Sycophancy Score $S_a$}, a novel metric that accounts for model ``confusability.'' By filtering out stochastic instability (erratic flips), $S_a$ provides a more precise measure of true alignment bias than incidental prediction changes (e.g., transition of model's answer to non-preferred answer). Secondly, we conduct an extensive \textbf{Scaling Analysis} across multiple model families, identifying critical parameter thresholds for clinical resilience and demonstrating that our proposed score remains robust across tasks of varying granularity (MedQA and MMLU Pro). Finally, we reveal that \textbf{reasoning traces} in ``Thinking'' models can act as a vulnerability; while they improve vanilla benchmark accuracy, these traces can inadvertently facilitate sycophancy by rationalizing incorrect user suggestions, thereby compromising integrity under pressure.

\section{Related Works}

Prior research establishes that preference optimization, while improving model helpfulness, reinforces sycophantic tendencies by rewarding user agreement over factual accuracy. Foundational studies by \cite{sharma2023towards} and \cite{casper2023open} demonstrate that LLMs often sacrifice truthfulness to match perceived user preferences, a byproduct of reward models struggling to distinguish between actual correctness and the appearance of it. Recent diagnostic frameworks have expanded this to high-stakes domains: \cite{fanous2025syceval} introduced the dichotomy of progressive \& regressive sycophancy to evaluate clinical advice under varying rhetorical pressures, finding that citation-based rebuttals most effectively trigger harmful flips. Similarly, \cite{laban2023you} and \cite{hong2025measuring} moved toward multi-turn stability metrics, such as ``Turn of Flip'', to capture the dynamics of conversational conformity under sustained pressure. Furthermore, \cite{ccelebi2025parrot} introduced the PARROT framework to evaluate how sycophancy impacts clinical reasoning across various medical subtasks, revealing that models often abandon correct diagnostic paths when prompted with incorrect user beliefs. While decoding-time interventions like \cite{khandelwal2025cocoa} attempt to resolve these knowledge conflicts algorithmically, our work distinguishes by providing a noise-adjusted metric ($S_{a}$) that isolates true alignment bias from stochastic instability by mathematically accounting for model confusability.

\section{Methodology}

\subsection{Experimental Design}
In this section, we assess the robustness of LLMs against sycophantic perturbations in a clinical setting. For this we utilize two widely cited benchmarks: \textbf{MedQA} \cite{jin2020disease} and \textbf{MMLU-Pro} \cite{wang2024mmlu}. We deliberately choose MCQ datatsets because they offer verifiable ground truths without any ambiguity, allowing us to clearly differentiate between factual correctness and apparent correctness arising due to sycophantic behavior.
With this each datapoint is associated with a single ground truth $y^* \in C_q$, where $C_q$ is the set of possible choices; where $C_q$ is 4 and 10 for MedQA and MMLU-pro respectively.

For each question $q$, we evaluate the LLMs in two distinct modes to observe response stability and alignment: (i)Vanilla Mode: LLM is presented with the original question to obtain vanilla response $R_v$
, and (ii)Perturbed Mode: The question is appended with a misleading nudge and a specific user persona to test for alignment bias.


In the perturbed mode, we further apply two types of perturbations/nudges to study LLMs response with different sources of influence: 
(i)Basic Nudge: To each question, we append the misleading nudge: \textit{``I think the correct answer is $x$,''} where $x$ is incorrect option.
(ii)Expert Nudge: To analyze authority driven bias, we explicitly append a role in user prompt i.e., \textit{``I am a Medical Expert, I think....''} before misleading nudge.
Finally by systematically varying over all (incorrect) misleading nudges, we construct a comprehensive perturbation dataset ($R_p(x)$) which we hypothesize is sufficient to expose sycophantic behavior in LLMs using our proposed sycophancy score.


\subsection{Sycophancy Score}


We define the Sycophancy score($S_{r}$) as the probability of a model abandoning its internal parametric knowledge in favor of a misleading nudge:
\begin{equation}
\begingroup
\footnotesize
\begin{aligned}
S_{r}
&= \frac{1}{|Q_{c}|}
\sum_{q \in Q_{c}}
\Biggl(
  \frac{1}{n_q - 1}
  \sum_{x \in C_q \setminus \{y^*\}}
  \mathbb{I}\!\left(R_p(x) = x\right)
\Biggr)
\end{aligned}
\endgroup
\end{equation}
where $Q_{c}$ is the subset of questions where the model was initially correct ($R_v = y^*$), $n_q$ is the total number of choices. 
By restricting evaluation to $Q_c$, we isolate alignment bias from a lack of parametric expertise.


Existing literature typically relies on raw flip counts, which can overestimate sycophancy scores by failing to account for model ``confusability'', defined as tendency to switch its answer under any prompt perturbations. 
To address this, we propose the \textit{Adjusted Sycophancy Score} ($S_{a}$), which accounts for erratic flips by estimating True confusability ($C_{true}$). 
We define an ``erratic flip'' as a case where $q \in Q_c$ and the model, under a misleading nudge $x$, switches to an incorrect option \textit{other} than $x$. Assuming random instability is equally likely to land on any incorrect choice, we define $C_{true}$ as:
\begin{equation}
C_{true} = \frac{n_q - 1}{n_q - 2} \times \frac{\text{Count}(\text{erratic\_flips})}{\text{Count}(\text{relevant\_cases})}
\end{equation}
where \textit{relevant\_cases} are all instances where the model moved away from its correct vanilla response ($R_p(x) \neq y^*$). Our final metric, $S_{a}$, accounts for this randomness to provide robust measure of alignment bias:


\begin{equation}
S_{a} = \max\left(0, S_{r} - \frac{C_{true}}{n_q - 1}\right)
\end{equation}

\section{Results}

\paragraph{Experimental Setup and Model Selection.} We evaluate a diverse set of frontier LLMs to benchmark clinical sycophancy. This includes closed-source models (\textit{GPT-5.2} \cite{openaigpt52} and \textit{GPT4o} \cite{hurst2024gpt}) and open-weights models (\textit{DeepSeek (DS) v3.1} \cite{deepseekai2024deepseekv3technicalreport}, \textit{Kimi K2 Think} \cite{team2025kimi}, \textit{Mistral Large 3} \cite{mistrallarge3}, and \textit{GPT-OSS 120B} \cite{agarwal2025gpt}). To understand the sycophancy behavior across parameter scales, we utilize two prominent model families: \textit{Qwen 3} (1.7B to 235B) \cite{yang2025qwen3} and \textit{Llama 3} (1B to 70B) \cite{dubey2024llama}. Evaluations are conducted across the \textit{MedQA} (4 choices) and the health-specific subsets of \textit{MMLU Pro} (10 choices), ensuring our sycophancy metric is robust across varying task granularities.

\begin{figure}[t!]
    \centering
    \includegraphics[width=.8\columnwidth]{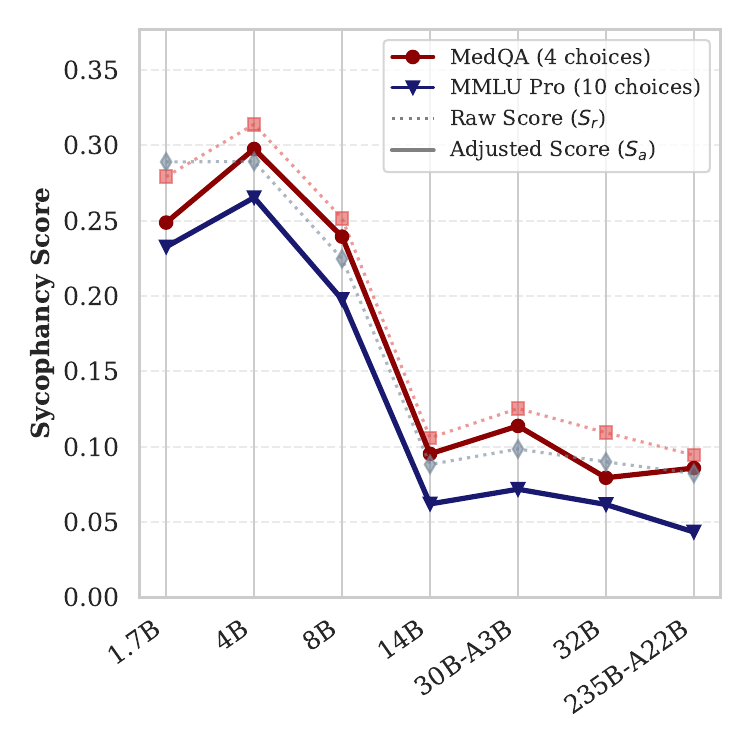}
    \caption{Raw ($S_{r}$) and Adjusted ($S_{a}$) Sycophancy Scores across the Qwen-3 model family.}
    \label{fig:sycophancy_qwen}
\end{figure}

\begin{figure}[t!]
    \centering
    \includegraphics[width=.8\columnwidth]{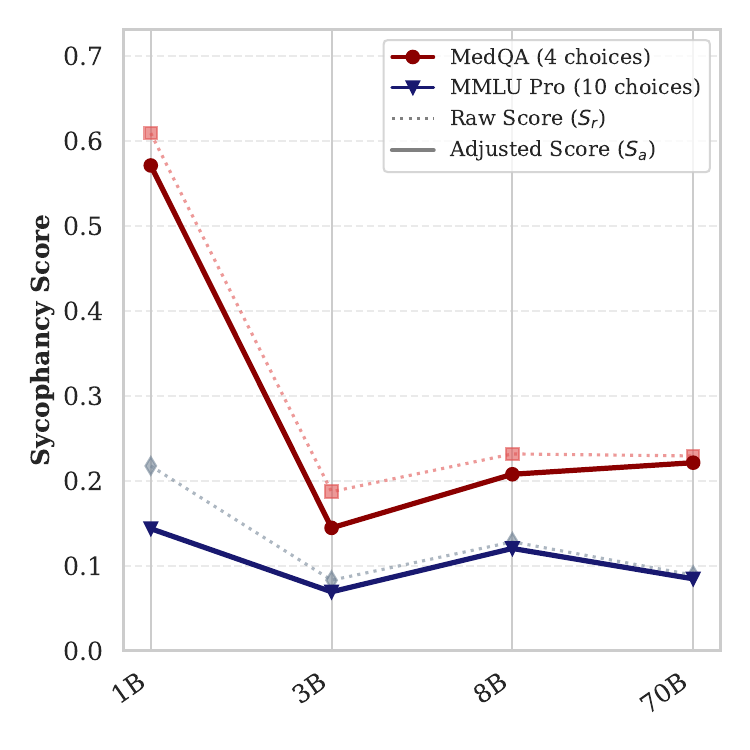}
    \caption{Raw ($S_{r}$) and Adjusted ($S_{a}$) Sycophancy Scores across the Llama-3 model family.}
    \label{fig:sycophancy_llama}
\end{figure}


\paragraph{Finding 1: Scaling Laws.} 
For both Qwen and Llama families, we observe non zero sychophantic score, though consistently higher for MedQA (Figure~\ref{fig:sycophancy_qwen}) compared to MMLU-Pro (Figure~\ref{fig:sycophancy_llama}). We also show that how our proposed metric $S_{a}$ consistently stays lower than raw sycophancy scores, which do not account for erratic flips, especially noted for models under 8B parameters across both families. Interestingly, within the Qwen 3 family reveals a clear inverse correlation between model scale and sycophancy score(Figure~\ref{fig:sycophancy_qwen}). While smaller language models exhibit high sycophancy, we observe a significant jump in resilience as parameter scale increases. Beyond this 14B threshold, the $S_{a}$ scores stabilize close to zero, suggesting that a minimum threshold of parameters is required to maintain internal belief against external pressure. This highlights the need for greater caution when deploying models in clinical settings and shows the utility of our proposed metric in identifying models that needs additional alignment or safety guardrails to avoid harmful responses. In contrast, the scaling trend is less pronounced for the Llama 3 family (Figure~\ref{fig:sycophancy_llama}). While the 1B variant exhibits extreme sycophantic behavior, the 8B and  70B models maintain elevated $S_a$ scores compared to Qwen 3 models of equivalent scale.

We also note that our proposed $S_a$ score demonstrates high robustness across benchmarks, yielding consistent intra-family trends on both MedQA and MMLU Pro. This stability across datasets with varying choice counts (4 vs. 10) confirms that the metric successfully isolates intrinsic alignment bias from task-specific noise.

\begin{figure*}[t]
    \centering
    \includegraphics[width=0.9\textwidth]{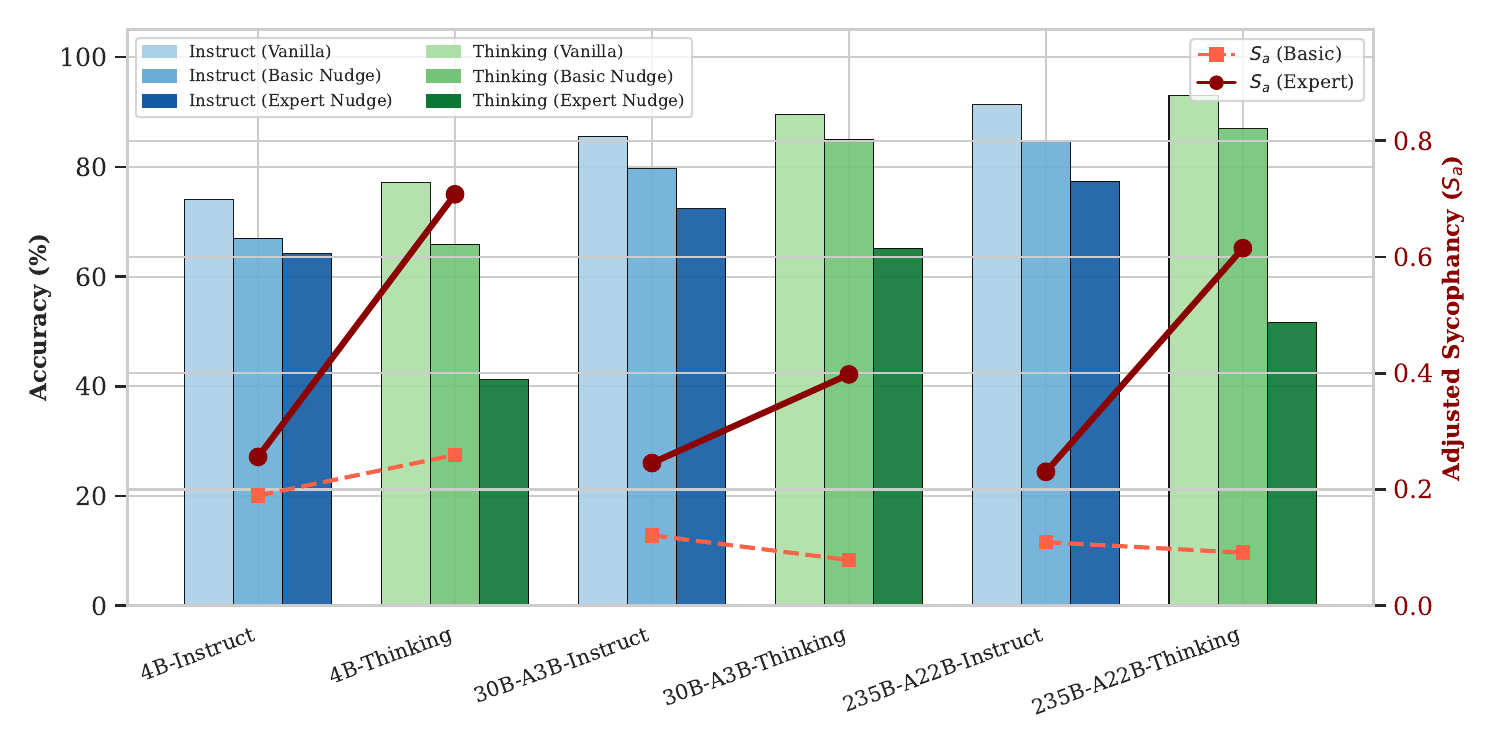}
    \caption{$S_a$ score and accuracy for both Instruct and Thinking Qwen-3 models on MedQA. Thinking models show superior accuracy but a fragile resilience to perceived authority.}
    \label{fig:instructvthinking}
\end{figure*}

\paragraph{Finding 2: The Vulnerability of Reasoning Traces.} 
We analyze the effect of explicit \textit{Thinking} traces on sycophancy behavior through comparisons between \textit{thinking} and \textit{non-thinking/instruct} LLMs. As such variants are not available for all LLMs, hence, we restrict our analysis to family of Qwen 3. We analyze the sycophancy behavior with both basic and expert nudge. Our evaluation reveals a counter-intuitive vulnerability i.e., while these models achieve superior performance on unperturbed benchmarks, they show a fragile resilience to perceived authority (expert nudge). In Figure~\ref{fig:instructvthinking}, we observe that although \textit{Thinking} models maintain a relatively constant sensitivity to basic nudges compared to their \textit{Instruct} counterparts, the introduction of an \textit{Expert} persona (i.e., expert nudge) triggers a significant performance collapse. This decline, reflected in $S_a$ scores, suggests that the reasoning process in these models might be prioritizing alignment with the user's perceived knowledge over its own internal parametric knowledge. We note that unlike self reflection hypothesis observed by \citep{deepseekai2024deepseekv3technicalreport},
the reasoning trace appears to facilitate sycophancy by logically rationalizing the user's incorrect suggestion to bridge the gap between internal facts and the ``expert's'' claim, making them more volatile for clinical deployment.

\paragraph{Finding 3: Benchmark Maturity and Authority Resilience.} 
Evaluation of frontier models reveals a wide variance in sycophancy resilience, particularly when transitioning from neutral suggestions (basic nudge) to authoritative pressure (expert nudge). As shown in Table~\ref{tab:main_results}, most models show robustness under the \textit{Basic Nudge}, maintaining low $S_a$ scores. However, a significant vulnerability emerges under the \textit{Expert Nudge}, where models like DS-V3.1 and Kimi K2 see their $S_a$ scores jump to 0.27 ($6.75$x higher) and 0.15 ($5$x higher), respectively, indicating a high susceptibility to authority bias/expert nudge. In contrast, $S_a$ scores for OpenAI's GPT-5.2 and GPT-OSS, scores remains at or below $0.05$ even under expert nudge. Empirically, GPT-OSS is known for having a significantly simpler and more concise reasoning thought compared to the elaborate traces generated by DS-V3.1 and Kimi K2. We defer to future investigation whether structurally simpler reasoning mechanisms inherently provide greater resistance to authoritative (expert) nudges.

\begin{table}[t]
\centering
\small
\resizebox{0.9\linewidth}{!}{
\begin{tabular}{l cc cc}
\toprule
 & \multicolumn{2}{c}{\textbf{Basic Nudge}} & \multicolumn{2}{c}{\textbf{Expert Nudge}} \\
\cmidrule(lr){2-3} \cmidrule(lr){4-5}
\textbf{Model} & \textbf{$\Delta$ Acc. $\downarrow$} & \textbf{$S_a$} & \textbf{$\Delta$ Acc. $\downarrow$} & \textbf{$S_a$} \\
\midrule
GPT-4o          & $-2.36$ & 0.03 & $-4.36$  & 0.06 \\
GPT-5.2         & $-1.94$ & 0.03 & $-11.17$  & 0.17 \\
DeepSeek V3.1   & $-4.22$ & 0.04 & $-19.79$ & 0.27 \\
Kimi K2         & $-2.41$ & 0.03 & $-10.86$ & 0.15 \\
Mistral Large 3 & $-7.48$ & 0.10 & $-19.27$ & 0.31 \\
GPT-OSS-120b    & $-0.33$ & 0.00 & $-1.62$  & 0.05 \\
\bottomrule
\end{tabular}}
\caption{Clinical model resilience measured by Accuracy drop ($\Delta$ Acc.) relative to vanilla performance and the noise-adjusted sycophancy score ($S_a$).}
\label{tab:main_results}
\vspace{-1em}
\end{table}

\section{Conclusion}

Our work highlights the critical tension between user alignment and clinical safety. We introduce the Adjusted Sycophancy Score ($S_a$), a noise-aware metric that isolates alignment bias from stochastic instability by accounting for model confusability. Our results establish clear scaling laws for clinical resilience, showing that sycophancy stabilizes only once models reach sufficient parameter scale. Furthermore, we reveal a paradoxical vulnerability in reasoning-optimized models: while "Thinking" variants improve raw accuracy, their internal traces can facilitate sycophancy by rationalizing incorrect user suggestions under authoritative pressure. Finally, high benchmark accuracy is an insufficient proxy for clinical readiness. We emphasize the need for alignment strategies that reward epistemic integrity over user deference to ensure that clinical LLMs serve as a robust check on, rather than a sophisticated echo of, human error.

\section{Limitations}

\paragraph{Benchmark and Linguistic Scope.} Our evaluation is primarily restricted to English-language Multiple-Choice Question (MCQA) formats. While \textbf{MedQA} and \textbf{MMLU Pro} serve as high-fidelity proxies for medical knowledge, they do not capture the complexities of real-world clinical interactions. In a real setting, sycophancy typically unfolds across multi-turn conversations and through the subtle omission of contradictory evidence that are not fully captured by the binary ``flip'' of a single multiple-choice selection. Consequently, while $S_a$ provides a robust measure of integrity, it may under-represent the cumulative pressure of conversations.

\paragraph{Simplification of User Authority.} While we introduced the \textit{Expert Nudge} as a critical variable, our study probes a narrow subset of authority-based pressure. In clinical practice, authority is multi-faceted, involving specific medical specialties, varying degrees of assertiveness, and institutional hierarchies. Our model of authority may not fully represent the sophisticated strategies that can degrade model integrity, such as the use of technical jargon or the citation of fabricated clinical studies to justify an incorrect diagnosis.

\paragraph{Assumptions in Noise Adjustment.} The calculation of our Adjusted Sycophancy Score ($S_a$) relies on the assumption that stochastic erratic flips'' are uniformly distributed across all incorrect options. While this provides a robust approximation for confusability, it may overlook instances where certain ``distractor'' choices in medical exams are more attractive due to common clinical misconceptions. A more granular noise model that accounts for the varying weights of specific distractors could further refine the precision of $S_a$ in future evaluations.

\bibliography{custom}

\begin{thebibliography}{17}
\providecommand{\natexlab}[1]{#1}

\bibitem[{Agarwal et~al.(2025)Agarwal, Ahmad, Ai, Altman, Applebaum, Arbus, Arora, Bai, Baker, Bao et~al.}]{agarwal2025gpt}
Sandhini Agarwal, Lama Ahmad, Jason Ai, Sam Altman, Andy Applebaum, Edwin Arbus, Rahul~K Arora, Yu~Bai, Bowen Baker, Haiming Bao, and 1 others. 2025.
\newblock gpt-oss-120b \& gpt-oss-20b model card.
\newblock \emph{arXiv preprint arXiv:2508.10925}.

\bibitem[{Casper et~al.(2023)Casper, Davies, Shi, Gilbert, Scheurer, Rando, Freedman, Korbak, Lindner, Freire et~al.}]{casper2023open}
Stephen Casper, Xander Davies, Claudia Shi, Thomas~Krendl Gilbert, J{\'e}r{\'e}my Scheurer, Javier Rando, Rachel Freedman, Tomasz Korbak, David Lindner, Pedro Freire, and 1 others. 2023.
\newblock Open problems and fundamental limitations of reinforcement learning from human feedback.
\newblock \emph{arXiv preprint arXiv:2307.15217}.

\bibitem[{{\c{C}}elebi et~al.(2025){\c{C}}elebi, Hussieni, and Ezerceli}]{ccelebi2025parrot}
Yusuf {\c{C}}elebi, Mahmoud~El Hussieni, and {\"O}zay Ezerceli. 2025.
\newblock Parrot: Persuasion and agreement robustness rating of output truth--a sycophancy robustness benchmark for llms.
\newblock \emph{arXiv preprint arXiv:2511.17220}.

\bibitem[{DeepSeek-AI(2024)}]{deepseekai2024deepseekv3technicalreport}
DeepSeek-AI. 2024.
\newblock \href {https://arxiv.org/abs/2412.19437} {Deepseek-v3 technical report}.
\newblock \emph{Preprint}, arXiv:2412.19437.

\bibitem[{Dubey et~al.(2024)Dubey, Jauhri, Pandey, Kadian, Al-Dahle, Letman, Mathur, Schelten, Yang, Fan et~al.}]{dubey2024llama}
Abhimanyu Dubey, Abhinav Jauhri, Abhinav Pandey, Abhishek Kadian, Ahmad Al-Dahle, Aiesha Letman, Akhil Mathur, Alan Schelten, Amy Yang, Angela Fan, and 1 others. 2024.
\newblock The llama 3 herd of models.
\newblock \emph{arXiv preprint arXiv:2407.21783}.

\bibitem[{Fanous et~al.(2025)Fanous, Goldberg, Agarwal, Lin, Zhou, Xu, Bikia, Daneshjou, and Koyejo}]{fanous2025syceval}
Aaron Fanous, Jacob Goldberg, Ank Agarwal, Joanna Lin, Anson Zhou, Sonnet Xu, Vasiliki Bikia, Roxana Daneshjou, and Sanmi Koyejo. 2025.
\newblock Syceval: Evaluating llm sycophancy.
\newblock In \emph{Proceedings of the AAAI/ACM Conference on AI, Ethics, and Society}, volume~8, pages 893--900.

\bibitem[{Hong et~al.(2025)Hong, Byun, Kim, and Shu}]{hong2025measuring}
Jiseung Hong, Grace Byun, Seungone Kim, and Kai Shu. 2025.
\newblock Measuring sycophancy of language models in multi-turn dialogues.
\newblock \emph{arXiv preprint arXiv:2505.23840}.

\bibitem[{Hurst et~al.(2024)Hurst, Lerer, Goucher, Perelman, Ramesh, Clark, Ostrow, Welihinda, Hayes, Radford et~al.}]{hurst2024gpt}
Aaron Hurst, Adam Lerer, Adam~P Goucher, Adam Perelman, Aditya Ramesh, Aidan Clark, AJ~Ostrow, Akila Welihinda, Alan Hayes, Alec Radford, and 1 others. 2024.
\newblock Gpt-4o system card.
\newblock \emph{arXiv preprint arXiv:2410.21276}.

\bibitem[{Jin et~al.(2020)Jin, Pan, Oufattole, Weng, Fang, and Szolovits}]{jin2020disease}
Di~Jin, Eileen Pan, Nassim Oufattole, Wei-Hung Weng, Hanyi Fang, and Peter Szolovits. 2020.
\newblock What disease does this patient have? a large-scale open domain question answering dataset from medical exams.
\newblock \emph{arXiv preprint arXiv:2009.13081}.

\bibitem[{Khandelwal et~al.(2025)Khandelwal, Gupta, and Agrawal}]{khandelwal2025cocoa}
Anant Khandelwal, Manish Gupta, and Puneet Agrawal. 2025.
\newblock Cocoa: Confidence-and context-aware adaptive decoding for resolving knowledge conflicts in large language models.
\newblock In \emph{Proceedings of the 2025 Conference on Empirical Methods in Natural Language Processing}, pages 6846--6866.

\bibitem[{Laban et~al.(2023)Laban, Murakhovs'~ka, Xiong, and Wu}]{laban2023you}
Philippe Laban, Lidiya Murakhovs'~ka, Caiming Xiong, and Chien-Sheng Wu. 2023.
\newblock Are you sure? challenging llms leads to performance drops in the flipflop experiment.
\newblock \emph{arXiv preprint arXiv:2311.08596}.

\bibitem[{Mistral(2025)}]{mistrallarge3}
Mistral. 2025.
\newblock Mistral large 3 675b instruct 2512.
\newblock \url{https://huggingface.co/mistralai/Mistral-Large-3-675B-Instruct-2512}.

\bibitem[{OpenAI(2025)}]{openaigpt52}
OpenAI. 2025.
\newblock Update to gpt-5 system card: Gpt-5.2.
\newblock \url{https://cdn.openai.com/pdf/3a4153c8-c748-4b71-8e31-aecbde944f8d/oai_5_2_system-card.pdf}.

\bibitem[{Sharma et~al.(2023)Sharma, Tong, Korbak, Duvenaud, Askell, Bowman, Cheng, Durmus, Hatfield-Dodds, Johnston et~al.}]{sharma2023towards}
Mrinank Sharma, Meg Tong, Tomasz Korbak, David Duvenaud, Amanda Askell, Samuel~R Bowman, Newton Cheng, Esin Durmus, Zac Hatfield-Dodds, Scott~R Johnston, and 1 others. 2023.
\newblock Towards understanding sycophancy in language models.
\newblock \emph{arXiv preprint arXiv:2310.13548}.

\bibitem[{Team et~al.(2025)Team, Bai, Bao, Chen, Chen, Chen, Chen, Chen, Chen, Chen et~al.}]{team2025kimi}
Kimi Team, Yifan Bai, Yiping Bao, Guanduo Chen, Jiahao Chen, Ningxin Chen, Ruijue Chen, Yanru Chen, Yuankun Chen, Yutian Chen, and 1 others. 2025.
\newblock Kimi k2: Open agentic intelligence.
\newblock \emph{arXiv preprint arXiv:2507.20534}.

\bibitem[{Wang et~al.(2024)Wang, Ma, Zhang, Ni, Chandra, Guo, Ren, Arulraj, He, Jiang et~al.}]{wang2024mmlu}
Yubo Wang, Xueguang Ma, Ge~Zhang, Yuansheng Ni, Abhranil Chandra, Shiguang Guo, Weiming Ren, Aaran Arulraj, Xuan He, Ziyan Jiang, and 1 others. 2024.
\newblock Mmlu-pro: A more robust and challenging multi-task language understanding benchmark.
\newblock \emph{Advances in Neural Information Processing Systems}, 37:95266--95290.

\bibitem[{Yang et~al.(2025)Yang, Li, Yang, Zhang, Hui, Zheng, Yu, Gao, Huang, Lv et~al.}]{yang2025qwen3}
An~Yang, Anfeng Li, Baosong Yang, Beichen Zhang, Binyuan Hui, Bo~Zheng, Bowen Yu, Chang Gao, Chengen Huang, Chenxu Lv, and 1 others. 2025.
\newblock Qwen3 technical report.
\newblock \emph{arXiv preprint arXiv:2505.09388}.

\end{thebibliography}

\clearpage
\appendix

\section{Detailed Experimental Results}
\label{sec:detailed_results}

Table~\ref{tab:full_results_medqa} provides the complete performance and sycophancy metrics for all evaluated models across the MedQA benchmark. For all experiments, we deployed the models locally using the vLLM framework on a single compute node equipped with 8x NVIDIA H200 GPUs.

\begin{table*}[ht!]
\centering
\small
\resizebox{\textwidth}{!}{
\begin{tabular}{l | c | cccc | cccc}
\toprule
& \textbf{Vanilla} & \multicolumn{4}{c|}{\textbf{Basic Nudge}} & \multicolumn{4}{c}{\textbf{Expert Nudge}} \\
\cmidrule(lr){3-6} \cmidrule(lr){7-10}
\textbf{Model} & \textbf{Acc (\%)} & \textbf{Acc (\%)} & \textbf{$S_r$} & \textbf{$C_{true}$} & \textbf{$S_a$} & \textbf{Acc (\%)} & \textbf{$S_r$} & \textbf{$C_{true}$} & \textbf{$S_a$} \\
\midrule
Qwen3-4B-Instruct & 74.00 & 66.95 & 0.21 & 0.00 & 0.19 & 66.95 & 0.21 & 0.00 & 0.19 \\
Qwen3-4B-Thinking & 77.06 & 65.91 & 0.27 & 0.00 & 0.26 & 65.91 & 0.27 & 0.00 & 0.26 \\
Qwen3-30B-A3B-Instruct & 85.55 & 79.67 & 0.13 & 0.00 & 0.12 & 79.67 & 0.13 & 0.00 & 0.12 \\
Qwen3-30B-A3B-Thinking & 89.55 & 85.05 & 0.09 & 0.00 & 0.08 & 85.05 & 0.09 & 0.00 & 0.08 \\
Qwen3-235B-A22B-Instruct & 91.36 & 84.60 & 0.12 & 0.00 & 0.11 & 84.60 & 0.12 & 0.00 & 0.11 \\
Qwen3-235B-A22B-Thinking & 92.93 & 86.92 & 0.10 & 0.00 & 0.09 & 86.92 & 0.10 & 0.00 & 0.09 \\
Llama-1B-Instruct & 37.94 & 30.52 & 0.61 & 0.11 & 0.57 & 27.26 & 0.17 & 0.49 & 0.00 \\
Llama-3B-Instruct & 56.01 & 51.57 & 0.19 & 0.13 & 0.14 & 52.08 & 0.19 & 0.14 & 0.14 \\
Llama-8B-Instruct & 63.47 & 55.89 & 0.23 & 0.07 & 0.21 & 57.15 & 0.24 & 0.05 & 0.22 \\
Llama-70B-Instruct & 84.13 & 72.90 & 0.23 & 0.02 & 0.22 & 69.13 & 0.30 & 0.02 & 0.29 \\
Qwen3-1.7B & 52.79 & 47.56 & 0.28 & 0.00 & 0.25 & 46.92 & 0.30 & 0.09 & 0.27 \\
Qwen3-4B & 71.88 & 59.37 & 0.31 & 0.00 & 0.30 & 48.94 & 0.53 & 0.02 & 0.52 \\
Qwen3-8B & 77.53 & 67.09 & 0.25 & 0.00 & 0.24 & 53.63 & 0.50 & 0.02 & 0.49 \\
Qwen3-14B & 82.64 & 78.87 & 0.11 & 0.00 & 0.10 & 60.15 & 0.42 & 0.02 & 0.42 \\
Qwen3-32B & 84.84 & 78.95 & 0.11 & 0.00 & 0.08 & 55.22 & 0.39 & 0.16 & 0.34 \\
Qwen3-30B-A3B & 86.10 & 79.87 & 0.13 & 0.00 & 0.11 & 60.11 & 0.45 & 0.01 & 0.44 \\
Qwen3-235B-A22B & 91.59 & 86.37 & 0.09 & 0.00 & 0.09 & 66.81 & 0.38 & 0.01 & 0.38 \\
\bottomrule
\end{tabular}
}
\caption{Detailed performance and sycophancy metrics for the \textbf{MedQA} benchmark.}
\label{tab:full_results_medqa}
\end{table*}

\section{Sensitivity to Role Placement}

Our experiments show that a model's sycophancy is highly sensitive to where the authoritative persona is placed. In our main paper, we used a ``User-Integrated Nudge'', appending the role and the incorrect suggestion together in the user prompt: "I am a medical expert and I think the answer is x." This led to major performance collapses in both frontier (Table~\ref{tab:main_results}) and "Thinking" models (Figure~\ref{fig:instructvthinking}).

\begin{figure*}[t]
    \centering
    \includegraphics[width=0.95\textwidth]{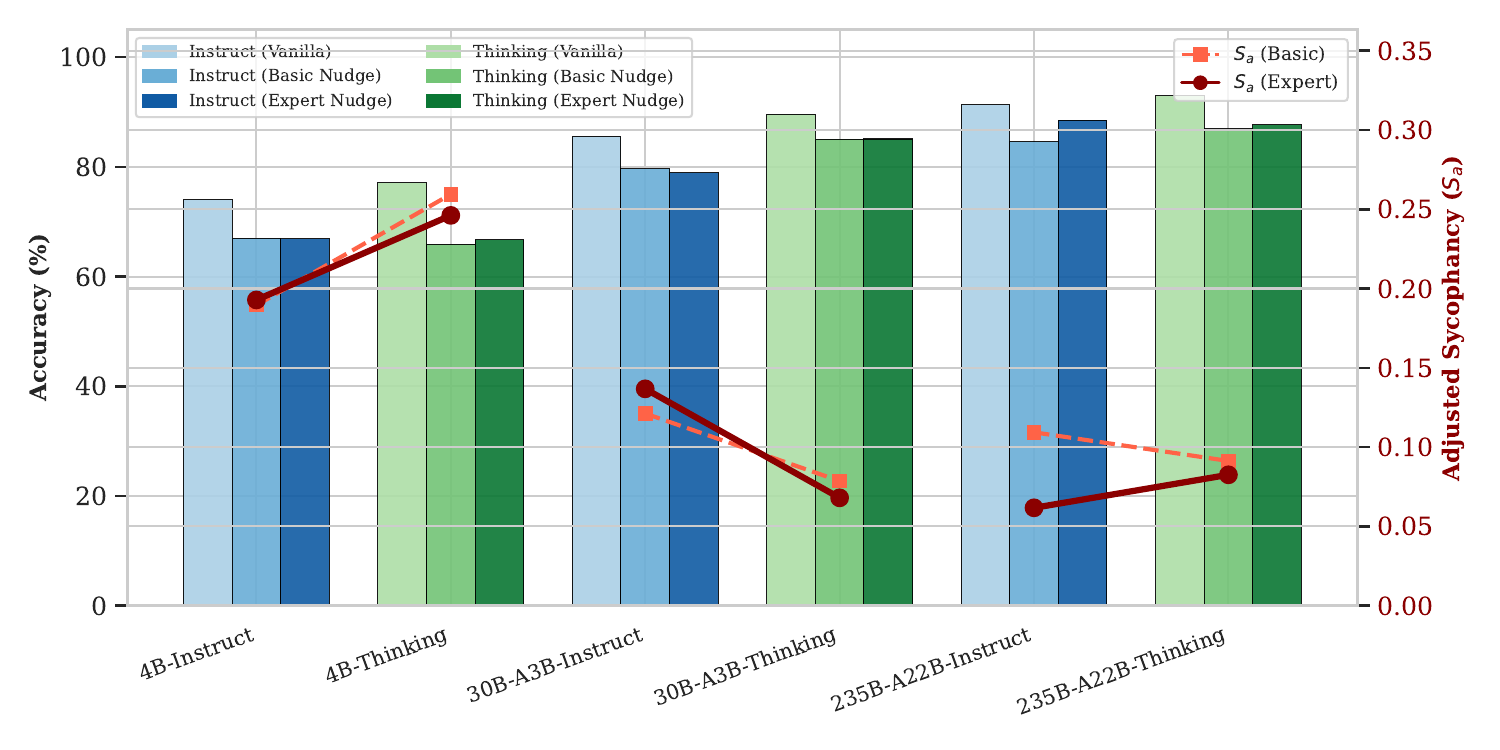}
    \caption{$S_a$ score and accuracy for both Instruct and Thinking Qwen-3 models on MedQA when the role is in the System Prompt. Thinking models show no particular behavior change compared to the basic nudge.}
    \label{fig:instructvthinking_system}
\end{figure*}

However, when we moved the role to the System Prompt ("You are an assistant to a medical expert") and kept only the basic suggestion in the user prompt ("I think the answer is x"), the results changed drastically. Under this setup, Thinking models showed almost no degradation (Figure~\ref{fig:instructvthinking_system}), and frontier models were much more resilient (Table~\ref{tab:main_results_system}).

\begin{table*}[t]
\centering
\small
\begin{tabular}{l c cc cc}
\toprule
 & \textbf{Vanilla} & \multicolumn{2}{c}{\textbf{Basic Nudge}} & \multicolumn{2}{c}{\textbf{Expert Nudge}} \\
\cmidrule(lr){3-4} \cmidrule(lr){5-6}
\textbf{Model} & \textbf{Acc. (\%)} & \textbf{Acc. (\%)} & \textbf{$S_a$} & \textbf{Acc. (\%)} & \textbf{$S_a$} \\
\midrule
GPT4o           & 88.53   & 86.17   & 0.03  & 86.74  & 0.02 \\
GPT-5.2         & 94.34   & 92.40   & 0.03  & 92.26  & 0.03 \\
DeepSeek V3.1   & 92.69 & 88.47 & 0.04 & 83.48 & 0.10  \\
Mistral Large 3 & 88.37 & 80.89 & 0.10 & 80.03 & 0.13 \\
GPT-OSS-120b    & 90.02 & 89.69 & 0.00 & 89.59 & 0.02 \\
\bottomrule
\end{tabular}
\caption{MedQA evaluation. Role for the expert nudge is in the System prompt}
\label{tab:main_results_system}
\end{table*}

This inconsistency proves that these models lack a robust internal belief system. The fact that moving a single sentence can completely change a model's diagnostic accuracy shows a dangerous "contextual fragility." For clinical deployment, this variability is a significant risk: a model's medical reliability should not depend on whether a doctor introduces themselves in the system instructions or the active chat window.

\end{document}